\pdfoutput=1
\documentclass[runningheads]{llncs}
\usepackage[T1]{fontenc}
\usepackage{graphicx}
\usepackage{url}
\usepackage{amsmath}
\usepackage{algorithm}
\usepackage{algpseudocode}
 \usepackage{booktabs}
 \usepackage{multirow}
 \usepackage{makecell}
 \usepackage{pifont}
 \usepackage{float}
 \usepackage{placeins}
\usepackage{listings}
\newif\ifblind
\blindfalse

\newif\ifarxiv
\arxivtrue

\begin{document}
\title{AdaNav: Query-Adaptive Multi-Granularity Navigation for Long Document Understanding}

\titlerunning{AdaNav: Adaptive Navigation for Long Document Understanding}
\ifblind
    \author{Anonymous Authors}
    \institute{Anonymous}
\else
    \author{Yiming Xu\inst{1}\orcidID{0000-0002-6579-9808} \and
            Eric López\inst{1}\orcidID{0009-0003-3229-5739} \and
            Artemis Llabrés\inst{1}\orcidID{0000-0002-6128-1796} \and
            Maximiliano Hormazábal\inst{1}\orcidID{0009-0003-3687-1924} \and
            Ernest Valveny\inst{1}\orcidID{0000-0002-0368-9697} \and
            Dimosthenis Karatzas\inst{1}\orcidID{0000-0001-8762-4454}}

    \authorrunning{Y. Xu et al.}
    \institute{Computer Vision Center, Universitat Autònoma de Barcelona, Barcelona, Spain
    \email{yiming@cvc.uab.es}\\}
\fi

\maketitle
\ifarxiv
\begingroup
\renewcommand\thefootnote{}
\footnotetext{This version of the contribution has been accepted for publication, after peer review, but is not the Version of Record and does not reflect post-acceptance improvements, or any corrections. The Version of Record is available online at: \url{https://doi.org/10.1007/978-3-032-36023-6_4}. Use of this Accepted Version is subject to the publisher's Accepted Manuscript terms of use \url{https://www.springernature.com/gp/open-research/policies/accepted-manuscript-terms}.}
\endgroup
\fi
\begin{abstract}
Long document understanding remains challenging for Vision-Language Models (VLMs) due to the need to locate relevant information across dozens of pages with diverse content types like text, tables, and figures. Existing embedding-based retrieval methods can miss relevant pages when the query–evidence relation is compositional. Embedding-free approaches avoid this by feeding all pages to VLMs, but incur high inference cost. Neither exploits the inherent hierarchical structure of documents for efficient navigation. In this paper, we propose AdaNav, an adaptive tree-based navigation framework for visual question answering in long documents. AdaNav constructs a multimodal document tree that hierarchically organizes sections, pages, and visual elements, enabling structured navigation without embedding retrieval models. We design the navigator to adapt the retrieval granularity to the question ensuring that content queries navigate from sections to pages in a coarse-to-fine manner. To support this, we build a multi-agent system with a question router and specialized navigators that differ in their action spaces, enabling traversal of the multimodal tree at the appropriate granularity. Experiments on MMLongBench-Doc demonstrate that AdaNav outperforms existing open-source VLM-based agent systems by over 5\% accuracy without requiring embedding models. Compared to embedding-free approaches, AdaNav retrieves fewer pages through tree-structured navigation. Our results show that heterogeneous tree representations with query-adaptive navigation offer an effective alternative to both paradigms for multimodal long document understanding.

\end{abstract}
\section{Introduction}

Visually rich long documents, such as financial reports, technical manuals, and scientific papers, typically span dozens of pages, with key evidence distributed across running text, tables, figures, and layout structure. Existing long-document QA pipelines \cite{mathew2021docvqa,tito2023hierarchical,ma2024mmlongbench} often fall into two extremes, making it difficult to achieve both high accuracy and efficiency. On the one hand, embedding-based retrieval~\cite{cho2025m3docvqa,jain2025simpledoc,lopez2025enhancing} uses page or segment embeddings for scalable search, but can miss relevant pages when the query–evidence relation is compositional (e.g., requiring multi-hop, cross-page, or cross-modal aggregation), causing downstream reasoning to fail due to missing evidence. On the other hand, embedding-free approaches~\cite{blau2024gram,tito2023hierarchical,zhu2025doclens} that feed all pages to a VLM avoid retrieval bias, but require processing the full document, leading to large token budgets and high inference cost; they also do not scale well to longer documents.

\begin{figure}[htbp]
\centering
\includegraphics[width=\textwidth]{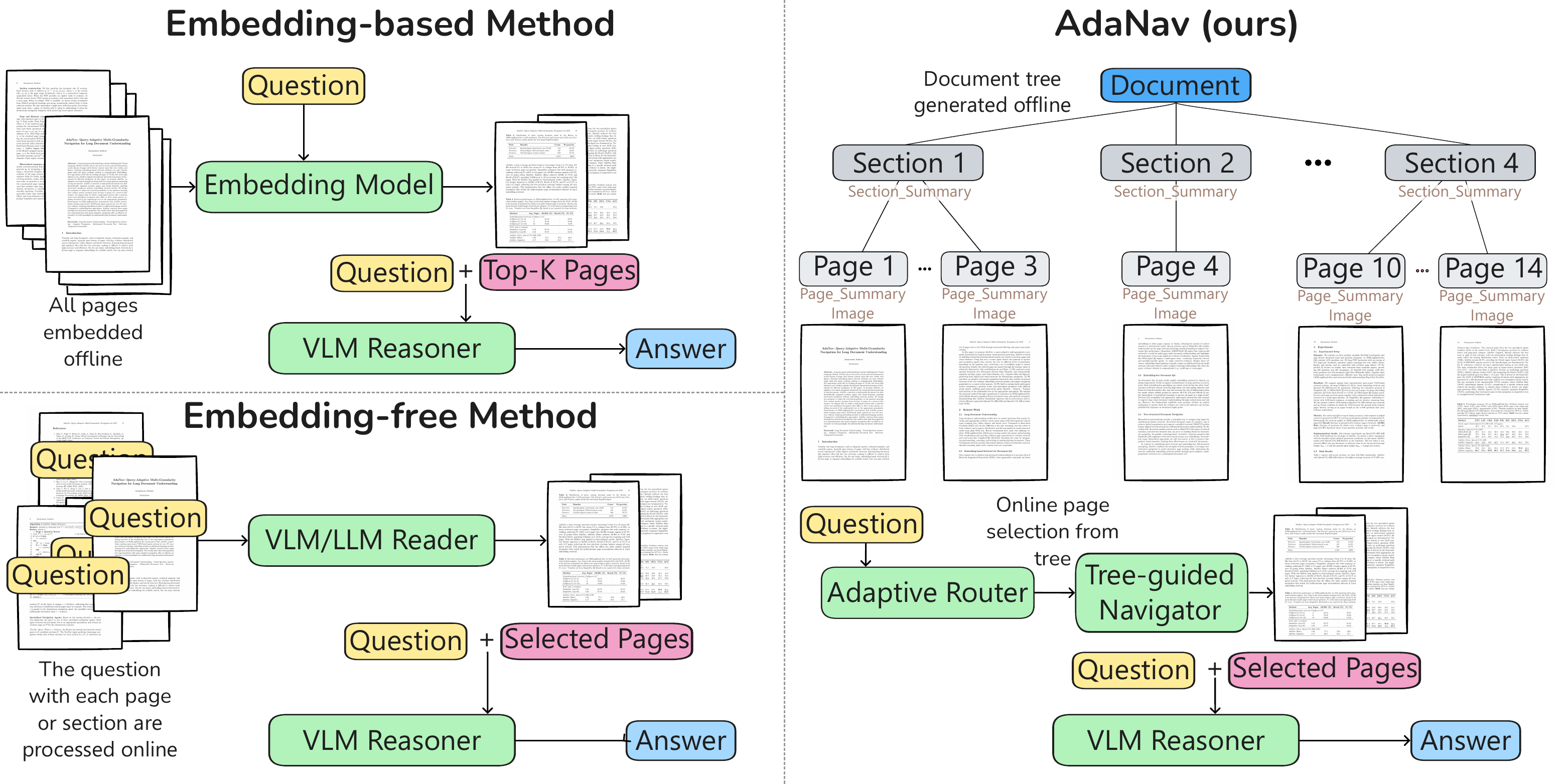}
\caption{Comparison between embedding-based methods (top-left), embedding-free methods (bottom-left) and AdaNav (right).}
\label{fig:method_comparison}
\end{figure}

We observe that documents are inherently hierarchical: sections group related content, pages contain localized information, and visual elements (tables, figures, charts) carry structured evidence. This structure naturally supports coarse-to-fine navigation, allowing a system to traverse from sections to pages to elements while filtering irrelevant branches early. Such navigation is particularly beneficial for cross-page and multi-hop questions, where evidence must be aggregated from multiple locations. This motivates a navigation-centric alternative that does not rely on extra embedding retrieval models, yet can structurally traverse a document and focus on only the necessary granularity for a given question.

Addressing this goal poses several challenges. First, long documents exhibit a natural hierarchy (Section → Page → Element), but choosing the right retrieval granularity is non-trivial: retrieving too coarsely introduces excessive irrelevant context, while retrieving too finely exponentially increases the search space. Building a multi-granularity, navigable representation and adapting the navigation granularity to the question remains an open challenge. Second, critical evidence often resides in table cells, chart annotations, legends, or localized visual regions. OCR-only representations lose layout and visual relations, while using page-level vision alone makes it difficult to align reliably the content of the page to question semantics; we need consistent grounding across section/page summaries and visual elements. Third, multi-page VLM inference is expensive, and errors can compound in long-document settings due to OCR noise, parsing errors, and cross-page dependencies. A practical system must reduce the number of pages sent to the VLM through structured filtering and query-type-aware routing.

In this paper, we propose AdaNav, Figure~\ref{fig:method_comparison}, a query-adaptive multi-granularity navigation framework for long-document visual question answering. AdaNav is based on building a hierarchical multimodal document tree based on sections, pages and visual elements. Using this tree, a router agent derives the question to specialized navigation agents that traverse the tree at different levels of granularity depending on the question type, retrieving a set of candidate pages to answer the question. Finally, the selected pages are passed through the reasoner agent to obtain the final answer. Our contributions are as follows.
(1) We construct a multimodal document tree (Document–Section–Page–Element) that hierarchically organizes sections, pages, and visual elements (e.g., cropped tables and figures), preserving both logical and visual structure for downstream navigation. (2) We introduce an adaptive tree-based navigation framework that enables structured traversal of the tree without embedding retrieval models and adapts navigation granularity in a coarse-to-fine manner. (3) We build a routing-based multi-agent system comprising a question router and specialized navigators with different action spaces, enabling query-type-aware paths (Specific / General / Normal) and traversal of the multimodal tree at the appropriate granularity. We validate the proposed method through comprehensive evaluation on MMLongBench-Doc and multiple datasets spanning diverse document types and question categories, demonstrating that AdaNav consistently improves end-to-end accuracy and retrieval efficiency using both Qwen3-VL-30B-A3B and Qwen2.5-VL-32B as backbone VLMs. Code and doctree data are available at \url{https://github.com/ym-xu/AdaNav}.

\section{Related Work}

\subsection{Long Document Understanding}

Long document understanding studies how to answer questions that require locating and aggregating evidence across many pages with heterogeneous content types (running text, tables, figures, and layout cues). Compared to short-form document QA\cite{masry2022chartqa,mathew2021docvqa}, the key difficulty is not only reasoning, but also where to look: evidence can be sparse, distributed, and often grounded in visual structures rather than plain OCR text. Recent benchmarks have made this challenge explicit. MMLongBench-Doc \cite{ma2024mmlongbench} focuses on long-context document understanding with visualizations and provides evidence page annotations to evaluate retrieval and end-to-end QA. LongDocURL \cite{deng2025longdocurl} further broadens the scope by integrating understanding, reasoning, and locating in multimodal long documents. These benchmarks motivate systems that jointly address evidence localization and multimodal reasoning under strict context and cost constraints.

\subsection{Embedding-based Retrieval for Document QA}

One common way to address long document understanding is to use some form of Retrieval Augmented Generation (RAG). These approaches commonly use dense embeddings to index pages, regions, or chunks, reducing the amount of context passed to a downstream reader. Recent systems such as SimpleDoc \cite{jain2025simpledoc} combine multi-modal cues for page retrieval and design careful prompting to improve document QA performance. Meanwhile, M3DOCRAG \cite{cho2025m3docvqa} argues that multi-modal retrieval is crucial for multi-page/multi-document understanding and highlights the limitations of text-only signals for evidence localization. Agentic frameworks like MDocAgent \cite{han2025mdocagent} employ a multi-agent team---consisting of general, critical, and modality-specific agents---to refine retrieved evidences. Despite these advances, embedding-based pipelines typically require an additional retrieval model and may still miss evidence under complex reasoning relationships, like when the query–evidence relation is compositional (e.g., multi-hop or cross-page).

\subsection{Embedding-free Document QA}

An alternative line of work avoids explicit embedding retrieval by directly applying long-context VLMs (or agentic orchestration) to large portions of a document. Such embedding-free paradigms can reduce retrieval bias but often “read” excessive irrelevant content, incurring high token cost and making inference inefficient for long documents; they may also increase the risk of hallucination when the model cannot reliably ground its answers. Hi-VT5~\cite{tito2023hierarchical} and GRAM~\cite{blau2024gram} utilize hierarchical or local-global strategies to process all pages in a single model. DocLens \cite{zhu2025doclens} exemplifies tool-augmented, multi-agent frameworks that attempt to improve long visual document understanding through orchestration and tool use.

\subsection{Tree-structured Document Navigation}

Hierarchical representations provide a principled way to navigate long inputs by exploiting document structure. Structural document maps, for instance, aim to preserve global organization and support controlled traversal; DMAP \cite{fu2026dmap} studies human-aligned structural maps for multimodal document understanding. On the tooling side, document parsing systems such as MinerU2.5 \cite{niu2025mineru2} expose structural metadata and detected elements that can serve as building blocks for hierarchical navigation. In parallel, reasoning-based ``vectorless'' retrieval narratives like PageIndex \cite{zhang2025pageindex} emphasize structured search actions over embeddings. Nevertheless, many hierarchical approaches are still text-centric or fail to preserve fine-grained visual semantics, limiting their effectiveness on visual-rich documents.

In contrast to embedding-based retrieval and embedding-free full-document prompting, AdaNav combines the strengths of both paradigms: it leverages tree-structured navigation to avoid exhaustive page reading, while eliminating the need for additional embedding retrieval models through query-adaptive, multi-granularity traversal over a multimodal document tree.

\section{Methodology}

\subsection{Overview}

AdaNav is a query-adaptive navigation framework for long-document visual question answering. The key insight is that different questions require different navigation granularities: some can be resolved by direct reference lookup (e.g., ``What does Figure 3 show?''), others need document-wide scanning (e.g., ``How many tables are there?''), and most benefit from coarse-to-fine section-to-page navigation (e.g., ``What is the proposed method?'').

Based on this observation, AdaNav operates in two stages: (1) an offline stage that parses the PDF and constructs a multimodal document tree capturing the hierarchy from sections to pages to visual elements; (2) an online stage that routes each question to a specialized navigator, retrieves relevant pages via tree traversal, and generates an answer with a VLM-based reasoner. We describe the tree construction in Section~\ref{sec:tree}, then detail the navigation pipeline in Section~\ref{sec:nav}.

\subsection{Multimodal Document Tree Construction}\label{sec:tree}

\begin{figure}[htbp]
\centering
\includegraphics[width=\textwidth]{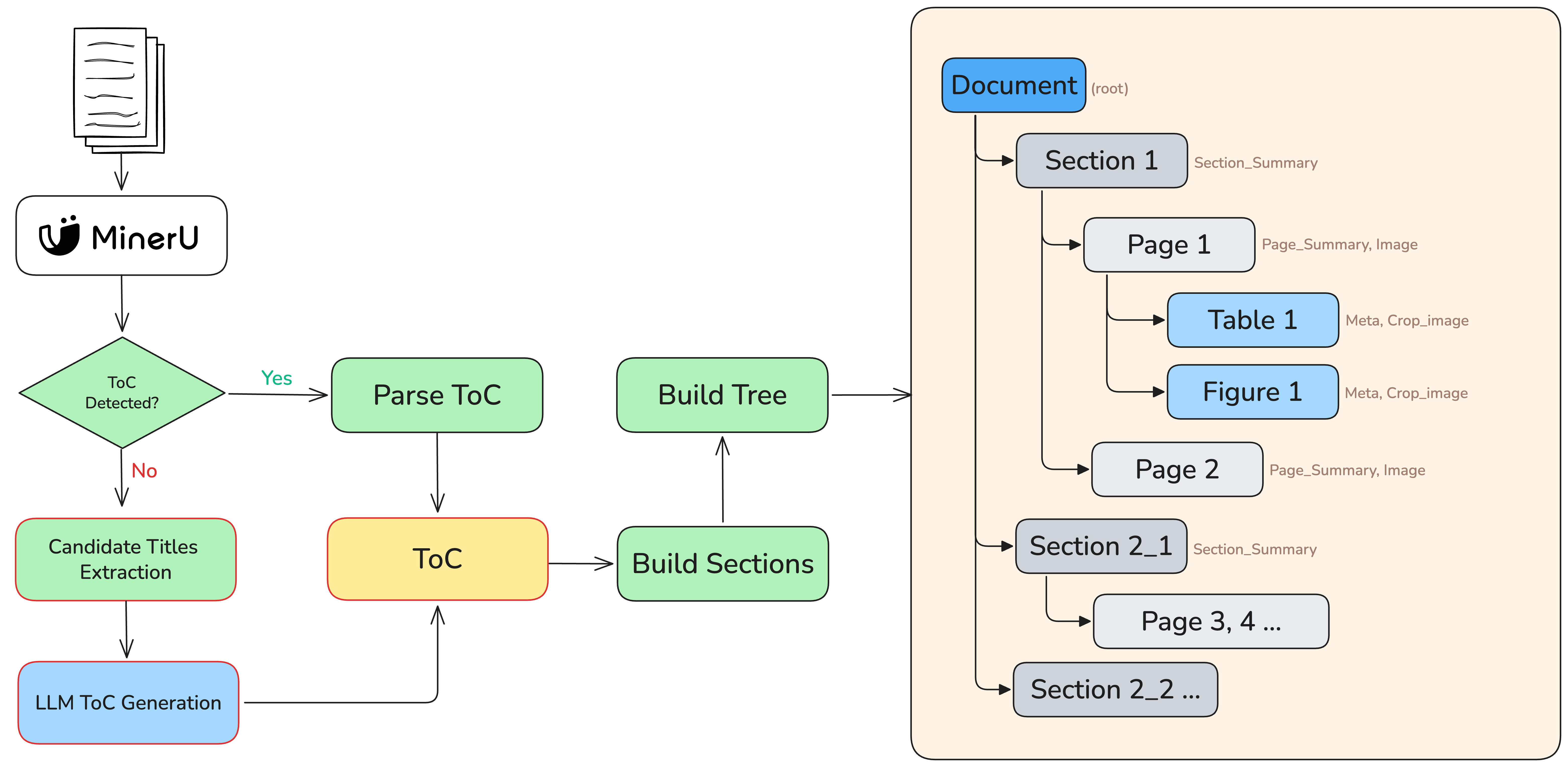}
\caption{Multimodal document tree construction. The tree organizes documents into four levels: Document (root), Section (with text summaries), Page (with screenshots), and Element (with cropped visual regions). This hierarchy enables coarse-to-fine navigation.}
\label{fig:tree}
\end{figure}

As illustrated in Figure~\ref{fig:tree}, we represent a document as a rooted tree \(\mathcal{T} = (D, \{S_i\}, \{P_j\}, \{E_k\})\) with four levels of nodes: a single Document node \(D\), a set of Section nodes \(\{S_1, \dots, S_M\}\), a set of Page nodes \(\{P_1, \dots, P_N\}\), and a set of Element nodes \(\{E_1, \dots, E_L\}\). The parent–child relations are \(D \to S_i \to P_j \to E_k\), forming a strict hierarchy. Each node carries a heterogeneous feature vector described below.

\textbf{Section construction.} We first partition the document into \(M\) sections. Each Section node is defined as \(S_i = (t_i, [a_i, b_i], \mathbf{s}_i)\), where \(t_i\) is the section title, \([a_i, b_i]\) is the page range (0-indexed), and \(\mathbf{s}_i\) is a section-level summary (populated later). When the PDF provides an explicit table of contents (ToC), we directly extract level-1 TOC entries as sections, each associated with a title and a start page. When no reliable TOC is available, we derive section boundaries from MinerU-predicted headings and group semantically related titles to form coherent sections. We only materialize a single layer of Section nodes. If a section spans more than \(\tau\) pages, we further split it using its subheadings to keep the downstream navigation budgeted while preserving local topical coherence.

\textbf{Page and Element construction.} Given the section structure, we assign each physical page to its parent Section according to page ranges, yielding \(N\) Page nodes.
Each Page node is defined as \(P_j = (I_j, x_j, \mathbf{p}_j, \{E_k\}_{k \in \mathcal{E}_j})\), where \(I_j\) is the rendered page image, \(x_j\) is the page-level textual content comprising the concatenated OCR text of all layout blocks extracted by MinerU, with each block annotated with its semantic type and bounding-box coordinates \([x_0, y_0, x_1, y_1]\), \(\mathbf{p}_j\) is a page-level summary, and \(\mathcal{E}_j\) indexes the element children of \(P_j\).

Each Element node is defined as \(E_k = (\text{type}_k,\, \text{bbox}_k,\, t_k,\, c_k,\, I_k^{\text{crop}})\), where \(\text{type}_k \in \{\texttt{table},\,\texttt{figure},\,\texttt{chart},\,\texttt{image},\,\texttt{diagram},\,\texttt{graph},\,\texttt{text},\,\texttt{page\_number}\}\) is the MinerU-assigned layout category, \(\text{bbox}_k\) is the bounding box on the source page, \(t_k\) is the block-level text content, \(c_k\) is the associated caption for figure and table elements, and \(I_k^{\text{crop}}\) is a cropped image present only for visual-category elements whose region exceeds a minimum size threshold.

\textbf{Hierarchical summary generation.} Beyond the structural hierarchy, we attach retrieval-oriented features at two granularities. For each Page node, we generate \(\mathbf{p}_j\) by prompting a VLM with the page image \(I_j\) and OCR text \(x_j\), using a structured template that elicits three components: (1) a free-text description of the page content; (2) textual and
visual evidence highlights, with separate fields for tables, figures, and charts; and (3) a categorised entity list covering numbers, names, dates, terms, and cross-page references. For each Section node, we generate \(\mathbf{s}_i\) by prompting a Large Language Model (LLM) with the concatenated page summaries \(\{\mathbf{p}_j\}_{j \in [a_i, b_i]}\), producing a structured index card that includes topic tags, evidence lists with page-level provenance, visual element inventories, a consolidated entity list. Crucially, the section card is intentionally designed to be queryable rather than abstractive, preserving specific values, figure/table identifiers, and cross-references verbatim to support precise retrieval matching.
Prompt templates are provided in the code repository.

\subsection{Query-Adaptive Navigation}\label{sec:nav}

\begin{figure}[htbp]
\centering
\includegraphics[width=\textwidth]{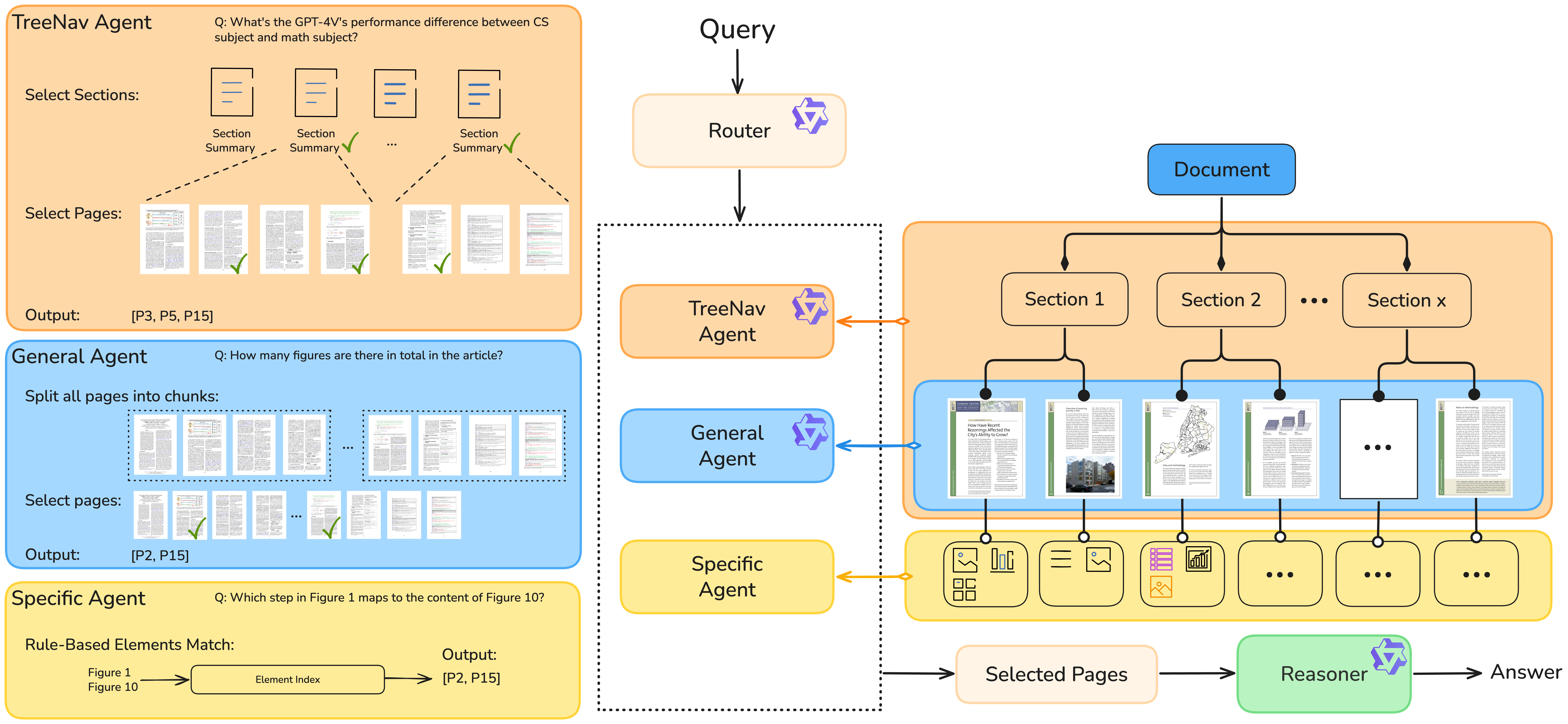}
\caption{AdaNav framework overview. \textbf{Left panel:} Workflows of three specialized agents. TreeNav selects top-K sections then navigates to relevant pages; General scans all pages in chunks for aggregation queries; Specific resolves element references via index lookup. \textbf{Right panel:} Query-adaptive routing dispatches each question to the appropriate agent based on its type, retrieving pages from the document tree at the optimal granularity before VLM reasoning.}
\label{fig:framework}
\end{figure}

As illustrated in Figure~\ref{fig:framework}, given a question and the prebuilt document tree, AdaNav answers through three stages: routing, navigation, and reasoning. Algorithm~\ref{alg:adanav}  summarizes the complete procedure.

\begin{algorithm}[t]
\caption{AdaNav Online Inference}\label{alg:adanav}
\begin{algorithmic}[1]
\Require question $q$; document tree $\mathcal{T} = (D, \{S_i\}_{i=1}^{M}, \{P_j\}_{j=0}^{N-1}, \{E_k\})$
\Ensure answer $\hat{a}$

\Statex \textbf{// Stage 1: Question Router}
\State $(\textit{refs},\; \mathcal{P}^*) \gets f_{\text{rule}}(q, \mathcal{T})$ \Comment{rule-based reference detection}
\If{$\textit{refs} \neq \emptyset$}
    \State $c \gets \textsc{specific}$ \Comment{direct lookup, zero LLM cost}
\Else
    \State $(\mathcal{S}^*,\; g) \gets f_{\text{llm}}\!\bigl(q,\; \{(t_i, \mathbf{s}_i)\}_{i=1}^{M}\bigr)$ \Comment{top-$K$ sections + global flag}
    \If{$g = 1$}
        \State $c \gets \textsc{general}$ \Comment{global query; all pages must be scanned}
    \Else
        \State $c \gets \textsc{normal}$;\; $\hat{\mathcal{S}} \gets \mathcal{S}^*$ \Comment{router-selected sections}
    \EndIf
\EndIf

\Statex \textbf{// Stage 2: Specialized Navigation}
\If{$c = \textsc{specific}$}
    \State $\hat{\mathcal{P}} \gets \mathcal{P}^*$ \Comment{resolved by index lookup, zero LLM cost}
\ElsIf{$c = \textsc{general}$}
    \State partition $\{P_0, \dots, P_{N-1}\}$ into overlapping chunks $\{\mathcal{C}_1, \dots, \mathcal{C}_L\}$
    \State $\hat{\mathcal{P}} \gets \emptyset$
    \For{each chunk $\mathcal{C}_l$} \Comment{all $N$ pages, token-budget chunking}
        \State $\hat{\mathcal{P}} \gets \hat{\mathcal{P}} \cup \pi_{\text{page}}(q,\; \mathcal{C}_l,\; B_{\text{gen}})$
    \EndFor
\Else \Comment{$c = \textsc{normal}$}
    \State $\hat{\mathcal{P}} \gets \emptyset$
    \For{each $S_i \in \hat{\mathcal{S}}$} \Comment{$K$ candidate sections}
        \State $\hat{\mathcal{P}} \gets \hat{\mathcal{P}} \cup \pi_{\text{page}}(q,\; S_i,\; B_{\text{nav}})$
    \EndFor
\EndIf

\Statex \textbf{// Stage 3: Multimodal Reasoning}
\State $\hat{a} \gets \text{VLM}\!\bigl(q,\;\{(I_j, x_j)\}_{j \in \hat{\mathcal{P}}}\bigr)$ \Comment{scratchpad reasoning over raw page content}

\State \Return $\hat{a}$
\end{algorithmic}
\end{algorithm}

\subsubsection{Router}\label{sec:router}

The router classifies each question into one of three categories:

\begin{itemize}
    \item \textbf{SPECIFIC}: the question contains explicit references such as ``page X'', ``figure Y'', or ``table Z'' (e.g., ``What does Table 3 show?''), which can be resolved by direct index lookup.
    \item \textbf{GENERAL}: the question requires document-wide coverage (e.g., ``How many figures are in this document?''), handled by scanning all pages.
    \item \textbf{NORMAL}: the question is a standard content query (e.g., ``What method does the paper propose?''), handled by coarse-to-fine navigation from candidate sections to pages.
\end{itemize}

Routing proceeds in two stages. In the first stage, a rule-based detector \(f_{\text{rule}}(q)\) pattern-matches explicit page, figure, and table references in the question; if any are found, the router assigns \(c = \textsc{specific}\) and terminates without invoking an LLM. In the second stage, an LLM-based classifier \(f_{\text{llm}}\!\bigl(q,\;\{(t_i, \mathbf{s}_i)\}_{i=1}^{M}\bigr)\) reads all \(M\) section titles and summaries jointly to determine whether the question can be localised to a subset of sections or requires full-document coverage. In the former case, it assigns \(c = \textsc{normal}\) and returns the top-\(K\) candidate sections \(\mathcal{S}^*\); in the latter, it assigns \(c = \textsc{general}\), indicating that targeted section selection is insufficient and all pages must be scanned. The routing decision \(c\) is passed to the downstream navigation agent; the candidate sections \(\mathcal{S}^*\) are additionally forwarded when \(c = \textsc{normal}\).

\subsubsection{Specialized Navigation Agents}\label{sec:agents}

Based on the routing decision \(c\), the system dispatches the query to one of three specialized navigation agents. Each agent traverses the document tree at an appropriate granularity and returns an evidence page set \(\hat{\mathcal{P}}\) for the downstream reasoner.

\paragraph{TreeNav Agent.}
When \(c = \textsc{normal}\), the Router has already narrowed the search space to \(K\) candidate sections \(\mathcal{S}^*\). The TreeNav Agent performs visual page navigation within each of these sections: for every section \(S_i \in \mathcal{S}^*\), it constructs an
interleaved multimodal sequence from the page images \(I_j\) and structured summaries \(\mathbf{p}_j\) of all pages \(P_j \in S_i\),
and presents this sequence to a VLM to identify the top-\(B_{\text{nav}}\) most relevant pages. For sections exceeding a length threshold \(\tau_{\text{pre}}\), a two-stage procedure is employed: an LLM first performs a coarse text-based pre-filter to reduce the candidate set, after which the VLM conducts fine-grained visual retrieval over the pruned pages. Results across all \(K\) sections are merged and deduplicated to form \(\hat{\mathcal{P}}\).

\paragraph{General Agent.}
When \(c = \textsc{general}\), the question demands document-wide coverage such as counting or aggregation across the entire document. The General Agent operates directly at the page level of the document tree, scanning all \(N\) page nodes without section-level filtering. As the VLM context is constrained by a token budget, the full page sequence is partitioned into overlapping chunks of size \(C\), where the overlap prevents evidence loss at chunk boundaries. For each chunk, the interleaved sequence of page images \(I_j\) and summaries \(\mathbf{p}_j\) is presented to the VLM for visual retrieval. The results from all chunks are merged via union to produce \(\hat{\mathcal{P}}\).

\paragraph{Specific Agent.}
When \(c = \textsc{specific}\), the question contains explicit references that can be resolved through direct index lookup. The Specific Agent first applies regular-expression patterns to extract reference targets from the question, such as page numbers, figure identifiers, and table identifiers. It then resolves each reference through a cascaded lookup over three offline-built indices: a logical-to-physical page number map, an element identifier index constructed from captions and block text matching ``Figure~\(X\)'' or ``Table~\(X\)'', and a page-level entity index for broader reference matching. The entire procedure incurs zero LLM cost, and the resolved page indices directly constitute \(\hat{\mathcal{P}}\).

\subsubsection{Reasoner}\label{sec:reasoner}

Once navigation produces the evidence set \(\hat{\mathcal{P}}\), a VLM generates the final answer:

\[\hat{a} = \text{VLM}\!\bigl(q,\;\{(I_j, x_j)\}_{j \in \hat{\mathcal{P}}}\bigr)\]

where \(I_j\) is the rendered page image and \(x_j\) is the OCR text of page \(P_j\). Notably, the structured retrieval indices built during the offline stage, namely page summaries \(\mathbf{p}_j\), section index cards \(\mathbf{s}_i\), and element-level crops \(I_k^{\text{crop}}\), are used exclusively for navigation and are not forwarded to the reasoner. The reasoner instead receives the raw page content directly, allowing the VLM to ground its answer in the original visual layout without interference from pre-computed abstractions. The prompt adopts a scratchpad structure, instructing the model to first identify key evidence from both modalities across the retrieved pages before synthesising the final answer.

\section{Experiments}

\subsection{Experimental Setup}

\subsubsection{Datasets.} We evaluate on three publicly available DocVQA benchmarks that span diverse document types and question categories. (1) MMLongBench-Doc \cite{ma2024mmlongbench} contains 1,073 questions over 135 long PDF documents with an average of 47.5 pages per document; questions require reasoning over text, tables, charts, figures, and layouts, and are annotated with evidence page indices. (2) PaperTab \cite{hui2024uda} focuses on tabular data extracted from academic papers, providing 393 questions over 307 documents. (3) FetaTab \cite{nan-etal-2022-fetaqa} contains 1,023 free-form questions over 878 documents built from Wikipedia tables. Together these benchmarks cover complementary difficulty axes: long multi-modal documents (MMLongBench-Doc) and structured table understanding (PaperTab, FetaTab).

\subsubsection{Baselines.} We compare against three representative open-source VLM-based retrieval systems, all using ColQwen-2.5 \cite{faysse2407colpali} for visual embedding retrieval and Qwen2.5-VL-32B-Instruct for generation, following the evaluation protocol of SimpleDoc \cite{jain2025simpledoc}. (1) M3DocRAG \cite{cho2025m3docvqa} retrieves the top-\(k\) pages via image embedding similarity and feeds them directly to a VLM. (2) MDocAgent \cite{han2025mdocagent} employs parallel text and image retrieval agents together with a dedicated critical-information extractor in a multi-agent pipeline. (3) SimpleDoc \cite{jain2025simpledoc} augments embedding retrieval with LLM-based summary re-ranking and an iterative reasoning loop. We also include a Direct VLM baseline (Qwen2.5-VL-32B without any retrieval) and an Oracle condition in which the VLM receives the ground-truth evidence pages directly, serving as an upper bound on the $n{=}847$ questions that carry evidence annotations.

\subsubsection{Metrics.} For end-to-end QA we report binary accuracy: each response is judged correct or incorrect by GPT-4.1 used as an automatic evaluator at temperature 0, following \cite{jain2025simpledoc}. For retrieval quality on MMLongBench-Doc we additionally report page-level \textbf{Recall} (fraction of ground-truth evidence pages retrieved), \textbf{All-Hit Rate} (fraction of questions for which every evidence page is retrieved), and page-level \textbf{F1} (harmonic mean of per-question precision and recall).

\subsubsection{Implementation details.} Our primary experiments use Qwen3-VL-30B-A3B as the VLM backbone for all stages of AdaNav. To enable a direct comparison with the baselines under identical generation conditions, we also report AdaNav results with Qwen2.5-VL-32B-Instruct as the backbone. The tree index is constructed offline once per document; at inference time we set the per-section page budget \(B_{\text{nav}} = 3\) and the general-agent budget \(B_{\text{gen}} = 2\) pages per section.

\subsection{Main Results}

Table~\ref{tab:main} reports end-to-end accuracy on three DocVQA benchmarks. AdaNav with Qwen3-VL-30B-A3B achieves the highest average accuracy, surpassing all open-source RAG baselines. On MMLongBench-Doc, AdaNav attains 65.73\%, a gain of +6.2 points over SimpleDoc and +10.4 points over MDocAgent, while retrieving significantly fewer pages. When evaluated on the answerable subset ($n{=}847$), AdaNav achieves 63.79\%, outperforming SimpleDoc by +4.2 points and MDocAgent by +8.5 points on this subset.

\begin{table}[!t]
\centering
\caption{Accuracy (\%) on three DocVQA benchmarks: MMLongBench-Doc (MMLB),
PaperTab, and FetaTab. \emph{Pg.\,Ret.} is the average number of pages
fed to the VLM per question. $^\dagger$Numbers are taken from
SimpleDoc~\cite{jain2025simpledoc}; all those methods use ColQwen-2.5 for visual
embedding retrieval and Qwen2.5-VL-32B-Instruct for generation.
$^\ddagger$Oracle is evaluated only on questions with annotated evidence pages
($n{=}847$ of $1{,}073$); the full pipeline covers all questions including
unanswerable ones. Best results are \textbf{bold}; second-best are
\underline{underlined}.}
\label{tab:main}
\setlength{\tabcolsep}{5pt}
\renewcommand{\arraystretch}{1.05}
\resizebox{\linewidth}{!}{%
\begin{tabular}{l c ccc c}
\toprule
\textbf{Method} & \textbf{Pg.\,Ret.} &
\textbf{MMLB} & \textbf{PaperTab} & \textbf{FetaTab} & \textbf{Avg.} \\
\midrule
\multicolumn{6}{l}{\textit{Direct VLM (no retrieval)}} \\
Qwen2.5-VL-32B$^\dagger$                        & --   & 22.18 &  7.12 & 16.14 & 15.15 \\
\midrule
\multicolumn{6}{l}{\textit{Oracle upper bound (GT evidence pages)$^\ddagger$}} \\
Qwen2.5-VL-32B + GT pages$^\dagger$             & --   & 67.94 &   --  &   --  &   --  \\
Qwen3-VL-30B-A3B + GT pages                     & --   & 67.65 &   --  &   --  &   --  \\
\midrule
\multicolumn{6}{l}{\textit{RAG Methods (Qwen2.5-VL-32B)$^\dagger$}} \\
M3DocRAG~\cite{cho2025m3docvqa}                  &  6   & 41.80 & 60.10 & 79.80 & 60.57 \\
MDocAgent~\cite{han2025mdocagent}                & 12   & 55.30 & \textbf{64.90} & \textbf{84.50} & \underline{68.23} \\
SimpleDoc~\cite{jain2025simpledoc}               & 3.2  & \underline{59.55} & \underline{64.38} & 80.31 & 68.08 \\
\midrule
\multicolumn{6}{l}{\textit{AdaNav (Ours)}} \\
AdaNav (Qwen3-VL-30B-A3B, answerable only)$^\ddagger$      & -- & 63.79 \phantom{0} & --  &   --  &   --  \\
AdaNav (Qwen2.5-VL-32B)                         & 4.17 & 59.46\phantom{0} & 61.06 & 82.89 &  67.80\\
AdaNav (Qwen3-VL-30B-A3B)                       & 4.17 & \textbf{65.73\phantom{0}}    &  63.61  & \underline{83.75} & \textbf{71.02} \\
\bottomrule
\end{tabular}
}
\end{table}

Across the three benchmarks, AdaNav shows complementary strengths and limitations relative to the baselines. On FetaTab, AdaNav (Qwen3) achieves 83.75\%, the second highest score, trailing MDocAgent (84.50\%) by less than one point; this small gap is consistent with observations in prior work~\cite{jain2025simpledoc} that MDocAgent's dedicated image retrieval agent confers a specific advantage on table-centric datasets. On PaperTab, AdaNav (Qwen3) achieves 63.61\%, slightly below SimpleDoc (64.38\%) and MDocAgent (64.90\%), indicating that structured tabular retrieval from academic papers remains a challenging domain where embedding-based retrieval complements tree navigation.

To enable a fair comparison under identical generation conditions, we also evaluate AdaNav with Qwen2.5-VL-32B-Instruct, the backbone used by all baselines. AdaNav (Qwen2.5) achieves 59.46\% on MMLongBench-Doc, which is on par with SimpleDoc (59.55\%) while operating without any pre-computed visual embeddings. On FetaTab, AdaNav (Qwen2.5) outperforms SimpleDoc by \(+2.6\) points (82.89\% vs.\ 80.31\%), confirming that the tree-structured navigation provides a reliable retrieval signal independent of the backbone. The improvement from Qwen2.5 to Qwen3 (\(+6.3\) points on MMLongBench-Doc, \(+2.5\) on FetaTab) reflects the stronger visual reasoning capability of Qwen3-VL-30B-A3B, particularly for multi-modal evidence types.

\subsection{Ablation Study}

\begin{table}[!t]
\centering
\caption{Ablation study on MMLongBench-Doc with Qwen3-VL-30B-A3B (evaluated
by GPT-4.1). Each row shows which agent handles each query category;
``TreeNav'' means the default tree-navigation path is used as a fallback.
Recall and F1 are page-level retrieval metrics (computed on $n{=}847$
questions with annotated evidence pages). ACC is overall accuracy on all
$1{,}073$ questions.}
\label{tab:ablation}
\setlength{\tabcolsep}{5pt}
\renewcommand{\arraystretch}{1.05}
\begin{tabular}{l ccc ccc}
\toprule
\multirow{2}{*}{\textbf{Variant}}
  & \multicolumn{3}{c}{\textbf{Query Routing}}
  & \multirow{2}{*}{\textbf{Recall}}
  & \multirow{2}{*}{\textbf{F1}}
  & \multirow{2}{*}{\textbf{ACC}} \\
\cmidrule(lr){2-4}
  & \textsc{Spec.} & \textsc{Gen.} & \textsc{Norm.} & & & \\
\midrule
AdaNav (Base)            & TreeNav  & TreeNav  & TreeNav  &  79.6  &  48.9  &   61.4  \\
w/o SpecificAgent        & TreeNav  & General  & TreeNav  &  83.9 &  56.2  &  63.4  \\
w/o GeneralAgent         & Specific & TreeNav  & TreeNav  &  83.2  &  53.7  &  62.8  \\
AdaNav (Agentic)         & Specific & General  & TreeNav  & \textbf{85.2} & \textbf{57.1} & \textbf{65.7} \\
\bottomrule
\end{tabular}
\end{table}

Table~\ref{tab:ablation} presents the ablation results. The base variant uses TreeNav for all queries, achieving 61.4\% accuracy. Adding the General Agent (+2.0\%) improves recall from 79.6\% to 83.9\% by scanning all pages in chunks for GENERAL queries. Adding the Specific Agent (+1.4\%) enables direct element lookup with zero LLM cost for SPECIFIC queries. The full AdaNav combines all three agents, achieving 65.7\% accuracy (+4.3\%), validating the effectiveness of query-adaptive routing.

\subsection{Analysis}

\paragraph{Query routing distribution.} Table~\ref{tab:routing} reports the routing decisions on MMLongBench-Doc, providing the interpretive context for all findings that follow. The vast majority of queries (76.7\%) follow the \textsc{Normal} path, reflecting that most document questions require structured coarse-to-fine navigation via the TreeNavAgent. The \textsc{Specific} path accounts for 10.5\% of queries (113 questions) involving explicit references to named pages, figures, or tables; these are resolved by the SpecificAgent through programmatic index lookup without any LLM call. The \textsc{General} path covers 12.8\% of queries (137 questions) requiring document-wide aggregation, handled by the GeneralAgent through a full-document scan. This distribution validates the multi-path design: each path serves a structurally distinct query type, and the SpecificAgent alone eliminates LLM overhead for over one in ten questions.

\begin{table}[htbp]
\centering
\caption{Distribution of query routing decisions made by the Router on
MMLongBench-Doc (1,073 questions). The \textsc{Specific} path incurs
zero LLM cost; \textsc{General} and \textsc{Normal} paths invoke the
text-based PageNavAgent.}
\label{tab:routing}
\setlength{\tabcolsep}{8pt}
\renewcommand{\arraystretch}{1.05}
\begin{tabular}{l l r r}
\toprule
\textbf{Path} & \textbf{Handler} & \textbf{Count} & \textbf{Proportion} \\
\midrule
\textsc{Specific} & SpecificAgent (rule-based, zero LLM) & 113 & 10.5\% \\
\textsc{General}  & GeneralAgent (full-document scan)    & 137 & 12.8\% \\
\textsc{Normal}   & TreeNavAgent (coarse-to-fine)        & 823 & 76.7\% \\
\midrule
Total             &                                      & 1,073 & 100\% \\
\bottomrule
\end{tabular}
\end{table}

\paragraph{Retrieval efficiency.} The routing design translates into distinctive retrieval characteristics, as shown in Table~\ref{tab:retrieval}. Pure embedding-based retrieval (ColQwen-2.5) exhibits a clear coverage--precision tension: increasing \(k\) from 2 to 10 raises All-Hit from 64.1\% to 83.6\% but causes F1 to collapse from 38.75\% to 18.38\%, as many irrelevant pages accumulate. SimpleDoc mitigates this with summary re-ranking, achieving F1~\(\approx\)62\% at 3.5 pages, but All-Hit remains capped at 67.4\%, over 12 points below AdaNav. AdaNav (Base) achieves All-Hit of 77.5\% and Recall of 83.2\%, matching ColQwen at \(k{=}10\) in coverage but requiring only 5.94 pages. AdaNav (Agentic) further
improves to All-Hit of 80.2\%, Recall of 85.5\%, and F1 of 57.1\% at only 4.17 pages, achieving the best precision--coverage balance among all compared systems. This demonstrates that the offline tree index enables targeted navigation that avoids the indiscriminate page accumulation inherent in top-\(k\) embedding retrieval.

\begin{table}[htbp]
\centering
\caption{Retrieval performance on MMLongBench-Doc ($n{=}847$ questions
with annotated evidence pages).
\emph{Avg.~Pages} is the mean number of pages fed to the VLM.
\emph{All-Hit} is the fraction of questions for which every gold
evidence page is retrieved.
\emph{Recall} is the mean fraction of gold pages retrieved per question.
\emph{F1} is the macro-averaged page-level F1 score.
$^\dagger$Numbers are from SimpleDoc~\cite{jain2025simpledoc};
Recall is not reported for those methods.}
\label{tab:retrieval}
\setlength{\tabcolsep}{5pt}
\renewcommand{\arraystretch}{1.05}
\begin{tabular}{l cccc}
\toprule
\textbf{Method}
  & \textbf{Avg.~Pages}
  & \textbf{All-Hit (\%)}
  & \textbf{Recall (\%)}
  & \textbf{F1 (\%)} \\
\midrule
\multicolumn{5}{l}{\textit{Embedding-based retrieval (ColQwen-2.5)$^\dagger$}} \\
ColQwen-2.5 ($k{=}2$)   &  2    & 64.12 & -- & 38.75 \\
ColQwen-2.5 ($k{=}6$)   &  6    & 76.42 & -- & 24.36 \\
ColQwen-2.5 ($k{=}10$)  & 10    & 83.60 & -- & 18.38 \\
\midrule
\multicolumn{5}{l}{\textit{RAG with re-ranking$^\dagger$}} \\
SimpleDoc (top-10)       &  3.19 & 65.72 & -- & 61.42 \\
SimpleDoc (top-30)       &  3.46 & 67.37 & -- & 62.22 \\
\midrule
\multicolumn{5}{l}{\textit{AdaNav (Ours, Qwen3-VL-30B-A3B)}} \\
AdaNav (Base)            &  5.94 & 77.5  & 79.6 & 48.9 \\
AdaNav (Agentic)         &  4.17 & 80.2  & 85.5 & 57.1 \\
\bottomrule
\end{tabular}
\end{table}

\paragraph{Evidence-type breakdown.} The retrieval gains from the two specialized agents manifest clearly at the answer level. Table~\ref{tab:breakdown} decomposes accuracy by evidence source and page-span category. AdaNav (Agentic, Qwen3) achieves the best score in eight of nine columns, with two particularly striking findings that directly reflect the routing distribution above. First, on table-centric questions (TAB), AdaNav attains 68.7\%, outperforming SimpleDoc (51.2\%) by +17.5 points and MDocAgent (56.1\%) by +12.6 points: the 10.5\% of SPECIFIC queries routed to the SpecificAgent are dominated by ``Table~X'' references, resolved via exact caption-index lookup at zero LLM cost. The same mechanism drives the large gain on figure-centric questions (FIG: 47.3\(\to\)57.7, +10.4~pts from Base to Agentic). Second, on multi-page questions (MUL), AdaNav Agentic reaches 54.9\%, outperforming SimpleDoc (39.2\%) by +15.7 points, with the largest absolute gain over Base (+11.4~pts). This is driven by the GeneralAgent: the 12.8\% of GENERAL queries requiring document-wide aggregation are answered by a full-document scan that uniform tree navigation cannot match. The one exception is the unanswerable (UNA) category, where AdaNav Base (76.0\%) outperforms Agentic (71.4\%): committing to a specific retrieval path reduces the model's tendency to abstain when evidence is absent. On single-page questions (SIN), AdaNav Agentic (71.0\%) narrowly surpasses SimpleDoc (70.8\%), confirming that tree-based coarse-to-fine navigation is competitive even on straightforward localization tasks.

\vspace{-5pt}
\begin{table}[htbp]
\centering
\caption{Per-category accuracy (\%) on MMLongBench-Doc.
Evidence sources: text (TXT), layout (LAY), chart (CHA), table (TAB),
figure (FIG); page count: single-page (SIN), multi-page (MUL),
unanswerable (UNA). $^\dagger$Baseline numbers are from
SimpleDoc~\cite{jain2025simpledoc} using Qwen2.5-VL-32B-Instruct.
Our results are evaluated by GPT-4.1.
Oracle provides GT evidence pages directly and has no UNA subset.
\textbf{Bold}: best per column (non-Oracle); \underline{underlined}: second best.}
\label{tab:breakdown}
\setlength{\tabcolsep}{4pt}
\renewcommand{\arraystretch}{1.05}
\small
\begin{tabular}{l ccccc ccc c}
\toprule
\textbf{Method}
  & \textbf{TXT} & \textbf{LAY} & \textbf{CHA} & \textbf{TAB} & \textbf{FIG}
  & \textbf{SIN} & \textbf{MUL} & \textbf{UNA} & \textbf{ACC} \\
\midrule
\multicolumn{10}{l}{\textit{Oracle upper bound (Qwen3-VL-30B-A3B, GT pages)}} \\
Oracle
  & 64.2 & 63.3 & 66.3 & 65.7 & 63.6 & 73.3 & 59.9 & --   & 67.7 \\
\midrule
\multicolumn{10}{l}{\textit{Baselines (Qwen2.5-VL-32B)$^\dagger$}} \\
M3DocRAG~\cite{cho2025m3docvqa}
  & 46.7 & 41.5 & 45.4 & 39.2 & 43.8 & 58.6 & 30.6 & 22.5 & 41.8 \\
MDocAgent~\cite{han2025mdocagent}
  & 57.5 & 50.0 & 54.7 & 56.1 & \underline{52.8} & 68.7 & 42.9 & 45.4 & 55.3 \\
SimpleDoc~\cite{jain2025simpledoc}
  & 59.9 & 51.3 & 54.9 & 51.2 & 51.2 & \underline{70.8} & 39.2 & 67.4 & 59.6 \\
\midrule
\multicolumn{10}{l}{\textit{AdaNav (Ours, Qwen2.5-VL-32B)}} \\
AdaNav (Agentic)
  & \underline{61.6} & \underline{54.6} & 50.6 & \underline{62.7} & 51.8
  & 66.7 & \underline{46.5} & 64.2 & 59.5 \\
\midrule
\multicolumn{10}{l}{\textit{AdaNav (Ours, Qwen3-VL-30B-A3B)}} \\
AdaNav (Base)
  & 59.8 & 52.1 & \underline{57.1} & 59.4 & 47.3
  & 67.7 & 43.5 & \textbf{76.0} & \underline{61.4} \\
AdaNav (Agentic)
  & \textbf{66.7} & \textbf{57.1} & \textbf{60.9} & \textbf{68.7} & \textbf{57.7}
  & \textbf{71.0} & \textbf{54.9} & \underline{71.4} & \textbf{65.7} \\
\bottomrule
\end{tabular}
\end{table}

\subsection{Case Study}
\begin{figure}[t]
\centering
\includegraphics[width=0.9\textwidth]{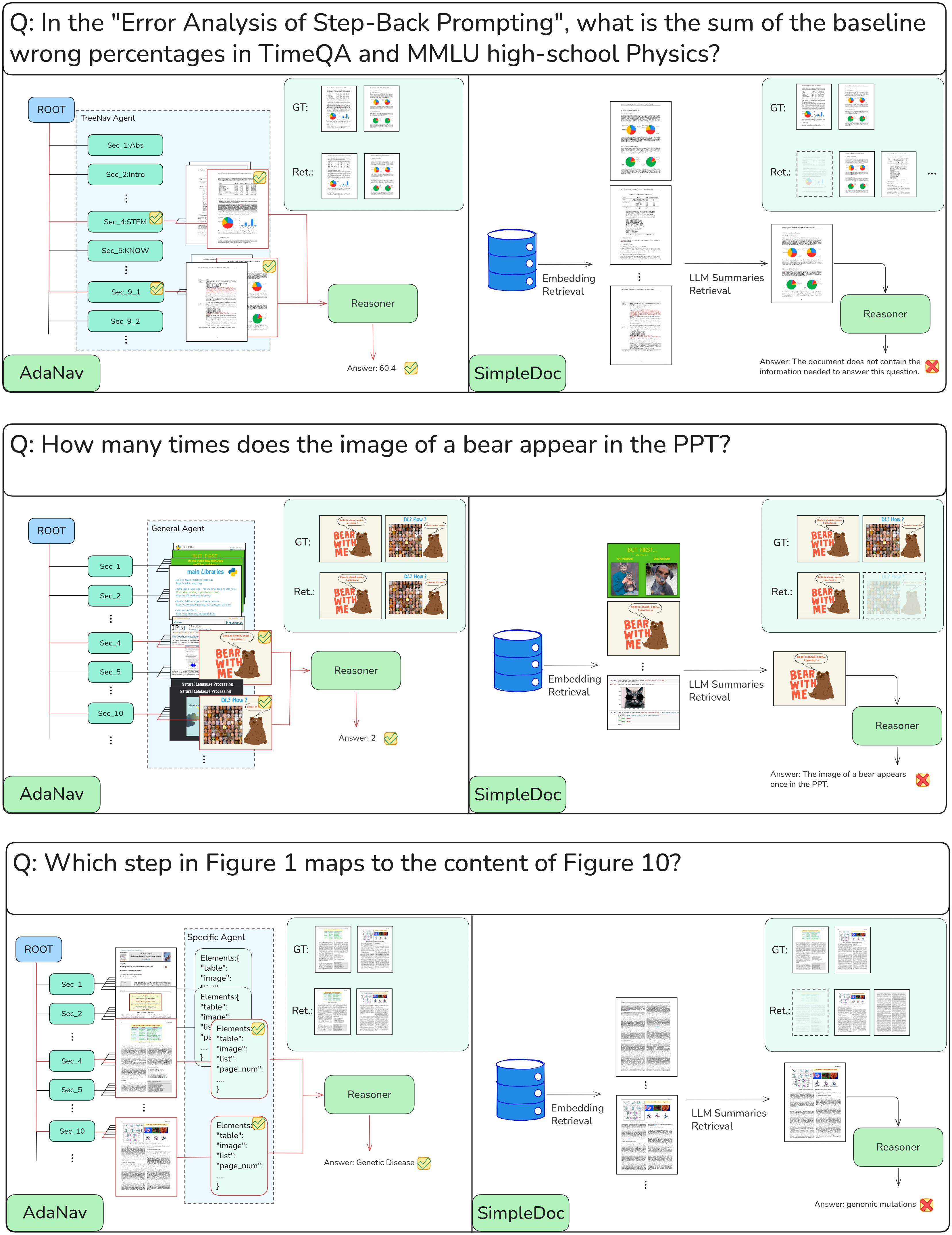}
\caption{Qualitative examples. A Normal question case where the TreeNav Agent is used (Top), a General question case where the General Agent is used (Middle), and a Specific Question where the Specific Agent is used (Bottom).}
\label{fig:qualitative-examples}
\end{figure}

Figure~\ref{fig:qualitative-examples} compares AdaNav with SimpleDoc on three representative question types. For the Normal question (Top), TreeNav selects relevant sections then navigates to pages containing the target table, while SimpleDoc's embedding retrieval fails to locate structural evidence. For the General question (Middle), the General Agent scans all pages in chunks to aggregate counts, whereas top-k retrieval misses occurrences. For the Specific question (Bottom), direct element index lookup retrieves the exact pages for Figure 1 and Figure 10 at zero LLM cost, while embedding similarity cannot reliably match figure references. These cases validate that different question types benefit from specialized navigation strategies.

\section{Limitations}

While AdaNav achieves strong performance on the evaluated benchmarks, several limitations remain.
The quality of the multimodal document tree depends directly on the reliability of the underlying parser. OCR errors in dense tables, incorrect heading detection in documents without a machine-readable table of contents, and missed figure captions introduce noise that propagates through every level of the tree; confidence-aware tree construction that can flag and recover from low-quality subtrees is a promising direction.
The current implementation constructs a tree per document. For book-length documents or cross-document settings, the Section level may need further decomposition, and inter-document linking is not yet supported.
Finally, the two-stage router is evaluated only on English documents in the MMLongBench-Doc domain; its generalization to domain-specific terminology or non-English text remains to be studied.

\section{Conclusion}
We have presented AdaNav, a query-adaptive multi-granularity navigation framework for long-document visual question answering. AdaNav offline constructs a multimodal document tree that hierarchically organizes sections, pages, and cropped visual elements, enabling structured navigation without embedding retrieval models. At inference time, a two-stage router classifies each question into one of three paths, and dispatches it to a specialized agent: the SpecificAgent resolves explicit element references through programmatic index lookup at zero LLM cost, the GeneralAgent performs a document-wide scan for global aggregation queries, and the TreeNavAgent executes coarse-to-fine section-to-page navigation for the remaining content questions. A shared GlobalReasonerAgent generates the final answer from the retrieved pages and visual elements.

Evaluated on MMLongBench-Doc, AdaNav (Agentic, Qwen3-VL-30B-A3B) achieves 65.7\% end-to-end accuracy, outperforming all open-source VLM-based agent baselines by a substantial margin while retrieving only 4.17 pages per query on average. A per-category breakdown confirms that each specialized agent yields gains on its intended evidence type, validating the multi-path design. These results suggest that structured offline indexing combined with query-adaptive online navigation constitutes an effective and practical alternative to both embedding-based and embedding-free paradigms for multimodal long document understanding.

\ifblind
\else

\section*{Acknowledgments}
This publication has been supported by the Consolidated Research Group 2021 SGR 01559 from the Research and University Department of the Catalan Government, and by project PID2023-146426NB-I00 funded by MCIU\allowbreak/AEI\allowbreak/\allowbreak10.13039\allowbreak/\allowbreak501100011033 and FEDER, UE. This work was also supported by the FI-STEP predoctoral grant program from the Department of Research and Universities of the Generalitat de Catalunya and co-financing by the European Social Fund Plus (2025STEP00060).

This work was also supported by the European Union's Horizon Europe research and innovation programme under grant agreements No 101214398 (ELLIOT) and No 101070617 (ELSA).

Funded by the European Union. Views and opinions expressed are however those of the author(s) only and do not necessarily reflect those of the European Union or the European Commission. Neither the European Union nor the European Commission can be held responsible for them.
\fi

\let\oldbibliography\thebibliography
\renewcommand{\thebibliography}[1]{\oldbibliography{#1}\setlength{\itemsep}{0pt}\setlength{\parsep}{0pt}}
\bibliographystyle{splncs04}
\bibliography{refs}

\clearpage
\appendix
\section{Prompt Templates}\label{app:prompts}

This appendix is not part of the published version of the paper. It lists the prompt templates used in the experiments; the same files are in the code repository (\texttt{prompts/}). Text in braces, such as \texttt{\{USER\_QUERY\}}, is a placeholder filled in at run time.

\subsection{Oracle}

In the Oracle condition (Tables~\ref{tab:main} and~\ref{tab:breakdown}), the VLM receives the ground-truth evidence pages as images, in the order of the annotation, followed by the prompt below. The Oracle does not use the AdaNav reasoner prompt.

\begin{lstlisting}
Answer the following question based on the provided document pages.

Question: {QUESTION}

Instructions:
- Analyze the page images carefully
- Provide a detailed answer based on the document content
- If the document does not contain enough information to answer, say "Not answerable"

Answer:
\end{lstlisting}

\subsection{AdaNav}

\paragraph{Page summary.} Produces the page summary \(\mathbf{p}_j\) from the page image \(I_j\) and its OCR text \(x_j\) (Section~\ref{sec:tree}).

\begin{lstlisting}
You are tasked with creating a comprehensive summary of a document page. Your summary should focus on extracting and describing the main content, tables, figures, and images present on the page.

Page OCR Text (if available):
{PAGE_TEXT}

## Instructions

Please analyze the page image and create a structured summary that captures:
1. The essential textual information from the page
2. Descriptions of any visual elements (tables, figures, images, charts)
3. Any particularly notable or unique information
4. **Specific numbers, names, dates, and key facts** that could answer questions
5. **Cross-page references** (e.g., "as shown in Table 3", "see Figure 5 on page 12")

## Output Format

You MUST output in this exact format with both tags:

<summary>
[4-7 sentences describing the page content. Be specific about what information is present.]

**Text Evidence:**
- [Key textual claims, findings, or statements with specific details]

**Visual Evidence:**
[For each visual element, include one of these tags:]
<table_summary>Table X: [title/description] - [key columns and notable values]</table_summary>
<figure_summary>Figure X: [title/description] - [what it shows and key values]</figure_summary>
<chart_summary>Chart X: [title/description] - [type of chart and key data points]</chart_summary>
<image_summary>[description of photos, maps, illustrations, diagrams - include key labels, locations, identifiable elements]</image_summary>
</summary>

<entities>
NUMBERS: [all numerical values with context, e.g., "95.2% accuracy", "$1.5M revenue", "25.6M parameters"]
NAMES: [people, companies, products, models, places, e.g., "ResNet-50", "ImageNet", "John Smith"]
DATES: [years, time periods, deadlines, e.g., "2023", "Q3 2024", "from 2019 to 2022"]
TERMS: [technical terms, acronyms, definitions, e.g., "CNN", "transformer", "backpropagation"]
REFERENCES: [cross-page/section references, e.g., "Table 3", "Figure 5", "Section 3.2", "as mentioned on page 12"]
</entities>

## Example

<summary>
This page (Page 15) introduces the concept of neural network architectures in Section 3.2. It discusses convolutional layers and their applications in image processing, noting that CNNs achieve 95.2% accuracy on ImageNet. The authors compare three popular architectures and highlight trade-offs between model size and performance. Results are compared with baseline methods shown in Table 1 on page 10.

**Text Evidence:**
- CNNs achieve 95.2% top-1 accuracy on ImageNet benchmark
- ResNet-50 provides best accuracy-to-parameter ratio
- Training was conducted for 100 epochs with batch size 256

**Visual Evidence:**
<table_summary>Table 1: Comparison of model parameters - ResNet-50 (25.6M params, 76.1% acc), VGG-16 (138M params, 71.5% acc), EfficientNet-B0 (5.3M params, 77.1% acc).</table_summary>

<figure_summary>Figure 2: Architecture diagram showing data flow through 5 convolutional layers and 3 pooling layers. Input size 224x224, output 1000 classes.</figure_summary>

<image_summary>Campus Map: Shows the university layout with labeled buildings (A-K), main roads including Zijing Road and Xuetang Road, dormitory areas, and key facilities. Location C marks the intersection of Zijing Road and Xuetang Road.</image_summary>
</summary>

<entities>
NUMBERS: 95.2% accuracy, 25.6M parameters, 138M parameters, 5.3M parameters, 76.1%, 71.5%, 77.1%, 224x224, 1000 classes, 100 epochs, batch size 256
NAMES: ResNet-50, VGG-16, EfficientNet-B0, ImageNet, NVIDIA A100, PyTorch
DATES: None
TERMS: CNN, convolutional layers, pooling layers, top-1 accuracy, backpropagation
REFERENCES: Table 1, Figure 2, Section 3.2, Table 1 on page 10
</entities>

## Key Focus Areas

Focus on information that would help answer questions such as:
- What are the key findings/numbers? (quantitative questions)
- What methods/approaches are described? (methodology questions)
- What comparisons are made? (comparison questions)
- What conclusions are drawn? (conclusion questions)
- What is shown in the figure/table/chart? (visual evidence questions)
- How many / What percentage / What is the total? (counting questions)

**CRITICAL: The <entities> tag is essential for retrieval. Structure entities by category (NUMBERS, NAMES, DATES, TERMS, REFERENCES) to enable precise matching.**
\end{lstlisting}

\paragraph{Section index card.} Produces the section card \(\mathbf{s}_i\) from the summaries of the pages in the section (Section~\ref{sec:tree}).

\begin{lstlisting}
You are tasked with creating a structured index card for a document section based on its page summaries.

Section Title: {SECTION_TITLE}
Pages: {START_PAGE} - {END_PAGE}

Page Summaries:
{PAGE_SUMMARIES}

## Instructions

Create a **structured index card** (NOT a prose summary) that preserves all queryable information from the pages. This card will be used for retrieval - users will search for specific terms, so DO NOT paraphrase or generalize.

## Output Format

You MUST output in this exact structured format:

<section_card>
SECTION: [section title] (Pages [start]-[end])

TOPICS: [comma-separated list of main topics covered]

TEXT_EVIDENCE:
- [Key textual claims, findings, or statements from this section]
- [Include page numbers where each claim appears]

VISUAL_EVIDENCE:
Tables:
- Table X (Page Y): [description with key columns and notable values]
- Table Z (Page W): [description]
(or "None" if no tables)

Figures:
- Figure X (Page Y): [description of what it shows]
- Chart Z (Page W): [type and key data points]
(or "None" if no figures)

Images:
- [Description of photos, maps, diagrams with key labels and locations]
(or "None" if no images)

KEY_ENTITIES:
NUMBERS: [all numerical values with context]
NAMES: [people, companies, products, models, datasets, places]
DATES: [years, time periods, deadlines]
TERMS: [technical terms, acronyms, definitions]
REFERENCES: [cross-page/section references, e.g., "see Table 3", "as shown in Figure 5"]

QUESTIONS_ANSWERABLE: [types of questions this section can answer, be specific]
</section_card>

## Example

<section_card>
SECTION: Experimental Results (Pages 12-18)

TOPICS: model comparison, ablation study, inference speed, accuracy benchmarks, training configuration

TEXT_EVIDENCE:
- ResNet-50 achieves 95.2% top-1 accuracy, outperforming all baselines (Page 13)
- Training was conducted for 100 epochs with batch size 256 on NVIDIA A100 GPUs (Page 14)
- Ablation study shows attention module contributes 15% improvement (Page 16)
- Authors conclude EfficientNet provides best accuracy-to-parameter ratio (Page 18)

VISUAL_EVIDENCE:
Tables:
- Table 3 (Page 13): Accuracy comparison across 5 models on ImageNet, COCO, Pascal VOC - ResNet-50 (95.2%), VGG-16 (91.3%), EfficientNet (94.8%)
- Table 4 (Page 15): Ablation study results - with attention (+15%), without (-3.2%)
- Table 5 (Page 17): Inference time comparison - ResNet-50 (2.3ms), VGG-16 (4.1ms), EfficientNet (1.8ms)

Figures:
- Figure 5 (Page 14): Training loss curves for all models over 100 epochs, showing convergence at epoch 80
- Figure 6 (Page 16): Accuracy vs. latency trade-off scatter plot, EfficientNet in optimal region
- Figure 7 (Page 18): Qualitative comparison of detection results on COCO images

Images:
- None

KEY_ENTITIES:
NUMBERS: 95.2% top-1 accuracy, 91.3%, 94.8%, 2.3ms inference time, 4.1ms, 1.8ms, 25.6M parameters, 15% improvement, batch size 256, 100 epochs, epoch 80
NAMES: ResNet-50, VGG-16, EfficientNet-B7, YOLO-v5, ImageNet, COCO, Pascal VOC, NVIDIA A100, PyTorch 1.9
DATES: None
TERMS: top-1 accuracy, ablation study, attention module, inference time, convergence
REFERENCES: Table 3, Table 4, Table 5, Figure 5, Figure 6, Figure 7, baseline methods in Section 2

QUESTIONS_ANSWERABLE: What is the accuracy of each model? Which model has the fastest inference time? What is the impact of the attention module? How many parameters does ResNet-50 have? What batch size was used for training? How many epochs were trained? Which model provides the best trade-off? What datasets were used for evaluation?
</section_card>

## Critical Rules

1. **DO NOT summarize into prose** - use the structured format above
2. **Separate TEXT_EVIDENCE from VISUAL_EVIDENCE** - users may ask "according to the table" or "in the text"
3. **Include page numbers** - helps locate evidence quickly
4. **Preserve ALL table/figure numbers** - these are critical retrieval hooks. Scan EVERY page summary for mentions of "Figure X", "Table X", "Chart X" and list ALL of them in VISUAL_EVIDENCE, even if the page summary only mentions them briefly. Missing a figure/table number means the retrieval system cannot find it.
5. **Preserve ALL specific numbers** - users often query exact values
6. **Structure KEY_ENTITIES by category** - enables precise matching for different question types
7. **Be exhaustive in QUESTIONS_ANSWERABLE** - list specific questions, not general categories
8. **Include cross-references** - note when content refers to other pages/sections
9. **Distinguish "contains" vs "references"** - In VISUAL_EVIDENCE, list figures/tables that are PRESENTED in this section. In REFERENCES, list figures/tables from other sections that are merely mentioned or cited.
\end{lstlisting}

\paragraph{Router.} The LLM stage \(f_{\text{llm}}\) of the router (Section~\ref{sec:router}). It reads the section titles and index cards and returns the top-\(K\) sections, or \texttt{ALL} for a question that needs the whole document (\textsc{general}).

\begin{lstlisting}
You are a document understanding agent tasked with identifying the most relevant section(s) for a given user query. You will be presented with summaries of each section in a document and a user query. Your task is to determine which section(s) should be examined in detail.

Document Overview:
{DOC_SUMMARY}

Section Summaries:
{SECTION_SUMMARIES}

User Query:
{USER_QUERY}

## Important Context
1. You are performing coarse-grained filtering based on section-level summaries.
2. The sections you select will be further analyzed at the page level.
3. Section summaries are high-level and may not capture all relevant details.
4. **Recall is critical** - missing a relevant section means potentially missing the answer entirely.
5. It's better to include a potentially relevant section than to exclude it.

## Instructions
1. Identify keywords, topics, and themes in the query that might appear in the document.
2. Select section(s) whose summaries suggest they contain information related to the query.
3. Be inclusive - if a section seems even somewhat related, include it.
4. Always select at least one section.
5. Order sections from most relevant to least relevant.
6. If the query combines several pieces of information (a comparison, several years, groups, items or pages, a list, or a count over a part of the document), select every section that may hold any one of the pieces - typically 2 to 5 sections.

## Global/Counting Queries (IMPORTANT)
Some queries require scanning the ENTIRE document. Use `ALL` ONLY for these cases:

**✓ TRUE Global Queries (output ALL):**
- "How many tables/figures/charts are in this paper?" → Need to scan entire document
- "How many pages contain tables?" → Need to check every page
- "List all countries mentioned in this report" → Need to scan entire document
- "How many times does 'X' appear in the document?" → Need full document scan
- "Count all bar charts in the report" → Need to examine all pages

**✗ NOT Global Queries (select specific sections):**
- "In Figure 3, how many icons are there?" → Only need the section with Figure 3
- "How many rows in Table 1?" → Only need the section with Table 1
- "What are the four business analytics activities in the Levels of Analytics chart?" → only need one chart
- "What comes between Reporting and Action in the flowchart?" → only need one flowchart
- "Which category has the most increase from 2005 to 2010?" → only need one chart/table
- "What does the map in the report show?" → only need the map page
- "How many organisations does this report introduce?" → can be found in a summary/overview section
- "How many chapter authors come from Columbia University?" → only need the author/contributor section
- "What are the last 4 chapters in the table of contents?" → only need the table of contents page
- "How many QR codes are in the Academics section?" → only need one section
- "Enumerate the media sources surveyed in this report" → only need the methodology section
- "What percentage of respondents favor X?" → only need one survey result section

**Key Distinction:**
- Global: Query needs to COUNT or LIST items ACROSS the entire document, where NO single section could contain the answer
- Not Global: Query asks about a specific figure/table/section, even if it involves counting within that item

## Decision Rule
1. Does the query require counting or listing items ACROSS the entire document, where NO single section above could contain the answer? → output **ALL**
2. Otherwise → select the relevant **section IDs**

## Query Analysis
Before selecting sections, briefly analyze:
1. What type of information does the query seek? (factual, numerical, conceptual, comparative)
2. Does it require full document scanning? If yes → ALL
3. Where in a typical document would this information appear?

## Output Format
<query_analysis>
[Brief analysis of the query type and expected location of answer]
</query_analysis>

<selected_sections>
[List the section IDs separated by commas, e.g., sec_01, sec_03, sec_05]
[OR output "ALL" if this is a global counting query requiring full document scan]
</selected_sections>

<reasoning>
[Brief explanation of why these sections were selected and what you expect to find]
</reasoning>
\end{lstlisting}

\paragraph{Page navigator.} The page selection \(\pi_{\text{page}}\) of the TreeNav and General agents (Section~\ref{sec:agents}). The page images, each with its summary, are inserted at \texttt{\{PAGES\}}.

\begin{lstlisting}
You are a document page navigator with vision capabilities.

## Task
Select the most relevant page(s) for answering the user's question by examining both the page images and their summaries.

## Context
Section: {SECTION_TITLE}{CHUNK_INFO}
Pages in this chunk: {NUM_PAGES}

## User Question
{USER_QUERY}

## Pages
Below you will see {NUM_PAGES} pages, each with:
1. A text summary describing the page content
2. The actual page image

Examine BOTH the summary AND the image carefully. The image may contain visual elements (tables, figures, charts) not captured in the summary.
{PAGES}

## Instructions
1. Examine each page's IMAGE carefully - look for tables, figures, charts, or specific text relevant to the question.
2. The summary may miss visual details, so prioritize what you SEE in the images.
3. Select ALL pages that might help answer: "{USER_QUERY}"
4. Be inclusive - if a page seems even somewhat related, include it.
5. Order from most relevant to least relevant.

## Output Format
<selected_pages>
[Page numbers separated by commas, e.g., 5, 8, 12]
</selected_pages>

<reasoning>
[Brief explanation of why these pages were selected based on what you observed]
</reasoning>
\end{lstlisting}

\paragraph{Text pre-filter.} For sections longer than \(\tau_{\text{pre}}\), the text-based pre-filter that reduces the candidate pages before the page navigator (Section~\ref{sec:agents}).

\begin{lstlisting}
You are pre-filtering pages for visual inspection. Select the most promising pages based on summaries.

Section: {SECTION_TITLE} (Pages {START_PAGE}-{END_PAGE})

Page Summaries:
{PAGE_SUMMARIES}

User Query:
{USER_QUERY}

## Instructions
1. This is a PRE-FILTERING step - be INCLUSIVE to avoid missing relevant pages.
2. Select up to {KEEP} pages that might be relevant.
3. If unsure, include the page (visual inspection will verify later).
4. Order from most likely relevant to least.

## Output Format
<selected_pages>
[Page numbers separated by commas, e.g., 1, 3, 5, 8]
</selected_pages>
\end{lstlisting}

\paragraph{General Agent refilter.} Narrows the union of the pages that the General Agent selects over all chunks.

\begin{lstlisting}
You are a precision filter for a document QA system. You have been given a set of candidate pages that were broadly selected as potentially relevant to a user's query. Your job is to filter out clearly irrelevant pages while preserving all pages that might contribute to the answer.

User Query:
{USER_QUERY}

Candidate Pages (pre-selected from the full document):
{PAGE_SUMMARIES}

## Step 1: Classify the Query Type

**COUNTING query**: The question asks "how many X", "count the X", or requires tallying items across pages.
Examples: "How many figures are in the document?", "How many charts in the first 10 pages?", "How many logos appear?"

**LOCATING query**: The question asks for specific facts, names, values, or lists that can be found on a few pages.
Examples: "What percentage of workers are in X?", "List all sections about Y", "What does the map show?"

## Step 2: Apply Filtering Rules

**If COUNTING**: You MUST keep ALL pages that contain or might contain an instance of the counted element. Missing even one page means an incorrect count. Only remove pages that clearly do NOT contain the element type being counted. You may keep up to {MAX_PAGES_COUNTING} pages.

**If LOCATING**: Be selective — remove pages that are only tangentially related. Keep the {MAX_PAGES} pages most likely to contain the answer.

## Output Format
<query_type>COUNTING or LOCATING</query_type>

<selected_pages>
[List the page numbers separated by commas, e.g., 3, 7, 15]
</selected_pages>

<reasoning>
[Brief explanation of filtering decisions]
</reasoning>
\end{lstlisting}

\paragraph{Reasoner.} The reasoner (Section~\ref{sec:reasoner}), with the retrieved page images attached. When the pages are insufficient, the reasoner may return an updated query instead of an answer; the updated query is navigated again and the new pages are added, for at most three reasoner calls.

\begin{lstlisting}
## ROLE
You are an expert AI assistant specializing in multimodal long document understanding. Your task is to carefully analyze the provided page images (which may contain text, figures, tables, and other content) and provide a precise answer to the user's question. A preceding agent has already retrieved these pages from the full document based on the question.

## Follow these instructions carefully:
- Core Objective: Your primary goal is to accurately and concisely answer the user's question based on the content of the provided document pages.
- Rules of numerical answers:
    - If the user asks for an absolute number (e.g., with questions like "How many...?"), you must first attempt to locate the number directly. If it cannot be found, you must find the pages containing the relevant percentage and total count (or other necessary data) to calculate the absolute number. If the calculated absolute number for discrete entities (e.g., people, companies, objects) is a decimal, you must round it to the nearest whole number.
    - If the user asks for a percentage (or proportion), you must first attempt to locate the percentage directly. If it cannot be found, you must find the absolute numbers of the subgroup and the total count (or other necessary data) to calculate the percentage.
    - If the user's question is ambiguous and does not explicitly specify a number or percentage (e.g., "What's the gap between...?"), you must default to providing the absolute value. If you can only find relative values (percentages) in the chart, you must make every effort to find a total number within the provided context to calculate the absolute value. Only return the relative value as a last resort if a total number cannot be found, and explain that you cannot find total number in this case.
- Page Numbering: Page numbers in the user's question typically refer to the number printed on the page image, not the page's index in the document file. For example, if a PDF's first page is the cover and the third page is the first page of content (labeled "Page 1"), a user's question about "page 1" refers to that third page. Similarly, when asked to provide a page number, you should return the printed page number from the image. Only return the page index if no number is printed on the page.
- Rule of faithfulness: Be faithful. If the provided pages do not contain sufficient information to answer the user's question, you should answer 'Not answerable'. For example, if the user asks for a man in green shirts, but there are only man in red shirts in the provided pages, you should answer 'Not answerable'; if the user asks for the boy playing badminton, but there are only boys playing football in the provided pages, you should answer 'Not answerable'; if the user asks for a certain year's data but the provided pages only contain data for other years, you should answer 'Not answerable'; if the user asks for the color of a certain object but the provided pages do not contain that object, you should answer 'Not answerable'. Similarly, if the question refers to a specific figure/table by number and describes certain content (e.g., "icons for different Disciplines in Figure 10"), but the actual figure with that number shows entirely different content (e.g., a photograph of a castle), you should answer 'Not answerable' — do not force a numerical answer like "0" based on absence.

## Question
<question>
{QUESTION}
</question>

## Retrieved Pages
<retrieved_pages>
{RETRIEVED_PAGE_NUMBERS}
</retrieved_pages>

## Extracted Text from Retrieved Pages
<page_text>
{PAGE_TEXT}
</page_text>

## Notes from Previous Iterations
<memory>
{MEMORY_NOTES}
</memory>

IMPORTANT: Images of the retrieved pages are attached after this prompt. The extracted text is provided in <page_text> above. You must analyze BOTH the visual images AND the extracted text to fully understand the document and answer the question accurately.

Use the following <scratchpad> tag to organize your thoughts and show your analysis process:
<scratchpad>
1. List the key elements you observe in the provided page images and extracted text
2. Identify any specific details that relate to the question
3. Make connections between the document information (from both images and text) and the question
4. Determine if the provided information is sufficient to answer the question
5. If insufficient, consider what specific content from other pages might contain the answer
</scratchpad>

Based on your analysis in the scratchpad, respond with exactly ONE of:

If the provided pages contain sufficient information:
<answer>
Your clear, concise, and direct answer. Just the fact, number, or short phrase.
</answer>

If you believe the answer exists elsewhere in the document but is not in the current pages:
<query_update>
A refined search query that preserves the original question's meaning but adds specific keywords to help retrieve the missing pages. The new query must directly lead to answering the original question.
</query_update>
<notes>
Concise summary of: (1) what you found so far, (2) what is still missing, (3) why the new query should find it.
</notes>

If the document likely does not contain the answer at all (use ONLY as last resort):
<not_answerable>
Not answerable
</not_answerable>

CRITICAL RULES:
- The <answer> tag must contain ONLY the direct answer. No explanations, no page references, no reasoning.
- Use <query_update> when you see partial evidence suggesting the answer is on other pages you haven't seen yet.
- Specifically: if the question refers to "Figure N" or "Table N" but the provided pages only contain figures/tables with different numbers, the referenced item is very likely on other pages — you MUST use <query_update> to search for it, not <not_answerable>.
- Use <not_answerable> ONLY when you are confident the entire document does not contain the answer. Prefer <query_update> over <not_answerable> when in doubt.
- Do NOT request pages already listed in <retrieved_pages>.
- Your response must include <scratchpad> followed by exactly one of: <answer>, <query_update>, or <not_answerable>.
\end{lstlisting}

\end{document}